\documentclass[sigconf,nonacm]{acmart}
\AtBeginDocument{%
  }
\usepackage{booktabs}
\usepackage{placeins}
\AtBeginDocument{\hypersetup{pdfauthor={Faith Ogunwoye, Didar Seyidov, Konstantin Gubernatorov, Dzmitry Tsetserukou}}}

\begin{document}
\title{JitterVLA: A Replay Benchmark for Motion Quality in VLA-Driven Human-Robot Manipulation}
\author{Faith Ogunwoye, Didar Seyidov, Konstantin Gubernatorov, Dzmitry Tsetserukou}
\affiliation{\institution{Intelligent Space Robotics Laboratory, Skolkovo Institute of Science and Technology}\city{Moscow}\country{Russia}}
\email{\{faith.ogunwoye, didar.seyidov, konstantin.gubernatorov, d.tsetserukou\}@skoltech.ru}

\begin{abstract}
Vision-language-action (VLA) policies issue sparse targets whose arrival at the robot controller is delayed and irregular, so a reasonable predicted action can still produce abrupt motion near people. Remedies are evaluated on different streams, delays and metrics. JitterVLA replays identical commands and arrival traces through alternative controllers, isolating the controller's effect. We apply it to synthetic reaches, hand trajectories from 20 people, three open VLA pipelines (838 episodes) and a physical UR3. \textbf{Interpolation lowered modeled p99 jerk for every source, by 83\% (615 to 106\,m/s$^3$) on human motion with 200\,ms delay, at a 2.0\,cm tracking cost}, and the effect replicated on 598 new episodes; jerk-limited generation lowered jerk further at a larger cost. \textbf{Entropy--discontinuity associations had opposite, replicated signs in two pipelines}, and entropy gating helped in one but not the other. \textbf{On the UR3, interpolation lowered accelerometer-derived jerk by 3.0--14.0\% in all six exploratory pairs}, while a position-derived metric did not resolve the contrast. JitterVLA supports screening before human-centred studies; code and data will be released.
\end{abstract}
\begin{teaserfigure}
\centering{\large\textbf{Project page:} \url{https://faithogunwoye.github.io/JitterVLA/}}\\[6pt]
\includegraphics[width=\textwidth]{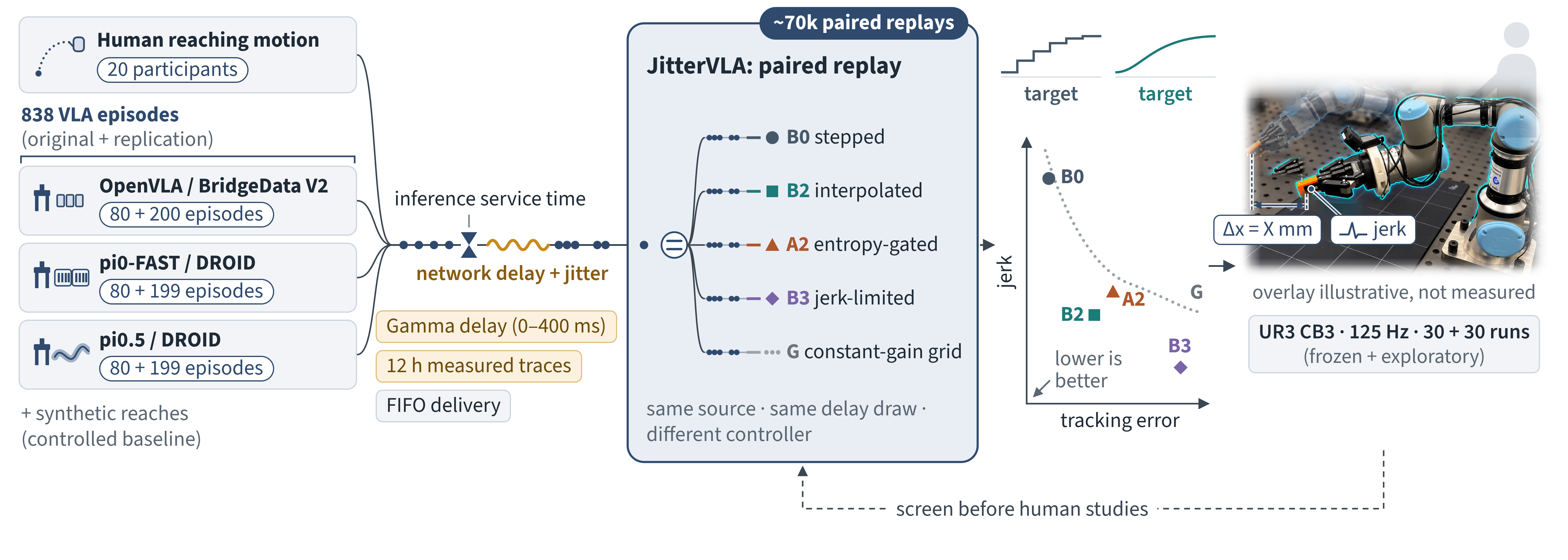}
\caption{JitterVLA. Commands from human motion and three VLA pipelines pass through inference and delivery timing; the \emph{same} commands and arrivals are replayed through alternative controllers, and the resulting jerk--tracking trade-off identifies which conditions warrant physical and human-centred evaluation.}
\Description{Flow from command sources through timing and paired replay to controller comparison and outputs.}
\label{fig:teaser}
\end{teaserfigure}
\maketitle
\renewcommand{\shortauthors}{Ogunwoye et al.}

\section{Introduction}
A person hands a manipulator an object, guides its target, or works beside it. The high-level policy produces a sensible action, but its targets arrive sparsely and at irregular times, so the low-level controller receives discontinuous goals. What the person experiences is not the predicted action but the motion the robot executes. VLA policies such as OpenVLA~\cite{kim2024openvla} run at a few hertz, one to two orders of magnitude below the robot's control loop~\cite{jiang2026fast}, and remote or shared inference adds network delay. Human-robot interaction (HRI) research links onset latency to teleoperation performance~\cite{rakita2020onset,khalil2026nonlinear} and robot motion to observers' responses and perceived safety~\cite{kulic2007physiological,rubagotti2022perceived}.

Timing is not only latency. A constant delay shifts every command, whereas \emph{timing jitter} (variation in delay and arrival spacing, including bunching) changes which targets the controller sees and when. Remedies exist at several layers: action chunking and asynchronous execution~\cite{zhao2023learning,black2026real,zhao2025vla,guo2026action,kim2025fine}, interpolation and jerk-limited trajectory generation such as Ruckig~\cite{berscheid2021jerk}, geometry-guided jerk reduction in VLA execution~\cite{yun2026groove}, and stiffness adaptation~\cite{ferraguti2013tank}. Each paper evaluates its remedy with its own inputs, delays, plant and metrics, so the remedies cannot be compared fairly. JitterVLA addresses that comparison problem. It evaluates the middle of the chain (command stream, timing, controller, executed motion) before candidate behaviours are shown to people. It does not replace human studies.

We ask four research questions (RQ1--RQ4). \textbf{RQ1 (timing):} how do timing irregularities change motion-quality proxies for identical command streams? \textbf{RQ2 (control):} how do stepped, interpolated, entropy-gated, constant-gain and jerk-limited strategies move the jerk--tracking trade-off? \textbf{RQ3 (cross-source):} do these relationships hold across synthetic, human and three VLA action representations, and on new episode draws? \textbf{RQ4 (physical):} are the contrasts observable on a real UR3 under a bounded protocol?

\textbf{Contributions.} (1) A \emph{paired replay} benchmark and protocol that holds the command source and delay realization fixed while varying the controller. (2) An evaluation that progresses from controlled synthetic reaches to 20-person human motion, three VLA pipelines with a planned replication on 598 new episodes, and measured network traces (Table~\ref{tab:design}). (3) \textbf{Findings: a consistent jerk--tracking trade-off for interpolation, representation-dependent uncertainty associations, and a failure of entropy gating on some inputs.} (4) A bounded UR3 check, including a null result for the planned position-derived metric. Code, the transformed VLA streams, the human windows and the traces will be released.

\textbf{Terms.} \emph{Latency} is delay; \emph{jitter} is its variation. \emph{Command discontinuity} is the spatial jump between successive targets or chunks. \emph{Jerk} is the benchmark plant's jerk proxy, or a sensor-derived estimate on the UR3. \emph{Tracking error} is the root-mean-square error (RMSE) against the undelayed reference. Jerk, tracking error, log dimensionless jerk (LDLJ) and spectral arc length (SPARC)~\cite{balasubramanian2015smoothness} are motion-quality \emph{proxies}; comfort, preference and safety were not measured.

\section{Benchmark design}
\textbf{Paired replay.} Within a comparison, every controller receives the same source (episode, person or synthetic reach) and the same delay realization (seed or trace offset). Differences are therefore attributable to the controller rather than to different stochastic inputs. Statistics are paired over the source unit: episodes, people or seeds. Intervals are 95\% unit bootstraps (5000 resamples), not time-sample intervals.

\textbf{Timing pipeline.} For VLAs, measured inference service times from one RTX~5090 feed a single-worker first-in, first-out (FIFO) queue. A network delay is then added: zero, Gamma-distributed with mean 0--400\,ms (shape 2), or sequential samples from 12\,h of measured lab-path traces: local area network (LAN), internet round trip, and a cold HTTPS request. Delivery is in order, and the newest delivered target wins. Separating the mean delay from its spread lets RQ1 test jitter as well as latency.

\textbf{Plant and controllers.} Delivered targets drive a generic three-axis point mass (3\,kg, critically damped impedance, 20\,ms force lag, 3000\,N/s force-rate cap, 1\,ms steps). It is \emph{not} a UR3 model; absolute jerk values are not hardware predictions. Controllers:
\begin{itemize}
\item \textbf{B0} holds each delivered target (stepped).
\item \textbf{B2} interpolates linearly over one command period.
\item \textbf{A2} adds entropy-gated stiffness to interpolation.
\item \textbf{B3} (Ruckig, jerk limit 25\,m/s$^3$) is an explicit jerk-limited trajectory generator.
\end{itemize}
Interpolation is the simple smoothing baseline; B3 tests whether gains survive a stronger motion-generation method. A \textbf{constant-gain grid} (8 fixed stiffness values, stepped and interpolated) asks whether gating beats the best \emph{attainable} constant gain at no worse tracking, rather than an obviously stiff baseline. We report 99th-percentile (p99) and root-mean-square (RMS) jerk and tracking RMSE, plus LDLJ and SPARC. LDLJ and SPARC are established dimensionless movement-smoothness measures, and they check that the conclusions do not depend on p99 jerk alone.

\textbf{Evidence hierarchy.} Protocols were frozen before data were seen: frozen analyses (v1.1, v1.3), a planned replication on new draws (v1.4), post-outcome robustness checks (v1.5: traces, LDLJ/SPARC, Ruckig, pi0.5~\cite{intelligence2025pi_}), and an exploratory hardware follow-up (v1.2a). We label results accordingly and never pool across these labels.

\begin{table}[t]
\caption{Experimental design. Every source is replayed under all controllers with a paired delay realization. N is original/new-draw replication. 69,988 paired replays in total: Gamma delays $10{\times}7$ (synthetic) $+\,20{\times}7{\times}10$ (human) $+\,838{\times}61$ (each VLA episode: zero delay once plus 6 means $\times$ 10 seeds) $=52{,}588$, and measured traces $3{\times}(10+10{\times}579)=17{,}400$ (synthetic, human, OpenVLA and pi0-FAST~\cite{pertsch2025fast} units).}
\label{tab:design}\centering\footnotesize\setlength{\tabcolsep}{2.6pt}
\resizebox{\columnwidth}{!}{%
\begin{tabular}{llll}
\toprule
Source & N & Command form & Uncertainty\\\midrule
Synthetic reaches & 10 & 5\,Hz targets & --\\
Human, 3D-ARM-Gaze~\cite{lento2024armgaze} & 20 people & 5\,Hz targets & --\\
OpenVLA / BridgeData~V2~\cite{walke2023bridgedata} & 80 / 200 & per-step tokens & token entropy\\
pi0-FAST / DROID~\cite{khazatsky2024droid} & 80 / 199 & 10-step chunks & chunk entropy\\
pi0.5 / DROID & 80 / 199 & 15-step flow chunks & none\\
UR3 (CB3 controller), 125\,Hz & 30 + 30 runs & replayed teleoperation motion & --
\\\bottomrule
\end{tabular}}
\\[2pt]\raggedright\footnotesize Timing: 7 Gamma means $\times$ 10 seeds, plus three measured traces; VLA service times measured. VLA actions are open-loop predictions on recorded observations, not commands those datasets' robots executed.
\end{table}

\section{Results}
\textbf{R1: interpolation trades jerk for tracking on every source (RQ1--3; frozen, replicated).} Interpolation lowered modeled p99 jerk and raised tracking error for synthetic, human and all three VLA sources (Table~\ref{tab:interp}). \textbf{The effect held on the new episode draws.} The reduction is not confined to synthetic or learned commands: it appears when replaying reaches from 20 people. \textbf{Timing irregularity, not mean delay, drove stepped-goal jerk.} In the frozen v1.0 sweep on synthetic reaches, pure fixed delays of 0, 200 and 400\,ms left stepped p99 jerk unchanged (535, 534 and 533\,m/s$^3$), whereas a 200\,ms delay spread raised it to 623. Under the three measured traces, the B2$-$B0 direction held for every source; for example, pi0-FAST gave $-368$, $-350$ and $-331$\,m/s$^3$ for LAN, internet and HTTPS. LDLJ and SPARC favoured B2 in every tested cell, and no interval crossed zero (post-outcome). The generic plant's force-rate cap accentuates stepped-goal differences, so these magnitudes are not hardware predictions.

\begin{table}[t]
\caption{Interpolation minus stepped (B2$-$B0) at zero added delay: modeled p99 jerk (m/s$^3$) and tracking RMSE (cm), original and new-draw replication. Negative jerk favours interpolation; positive RMSE is a tracking cost.}
\label{tab:interp}\centering\footnotesize\setlength{\tabcolsep}{4pt}
\begin{tabular}{lrrrr}
\toprule
& \multicolumn{2}{c}{Original} & \multicolumn{2}{c}{New draw}\\
Source & $\Delta$jerk & $\Delta$RMSE & $\Delta$jerk & $\Delta$RMSE\\\midrule
Synthetic (10) & $-504$ & $+2.39$ & -- & --\\
Human (20 people) & $-521$ & $+2.32$ & -- & --\\
OpenVLA (80 / 200) & $-202$ & $+0.78$ & $-193$ & $+0.77$\\
pi0-FAST (80 / 199) & $-371$ & $+0.28$ & $-343$ & $+0.26$\\
pi0.5 (80 / 199) & $-374$ & $+0.56$ & $-363$ & $+0.54$\\\bottomrule
\end{tabular}
\end{table}

\begin{figure}[t]
\centering\includegraphics[width=\columnwidth]{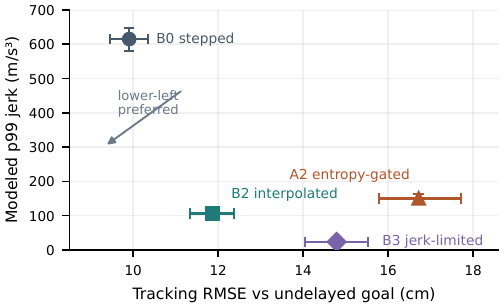}
\caption{Jerk--tracking frontier for hand trajectories from 20 people with 200\,ms mean Gamma delay. Seeds are averaged within each person; 95\% intervals resample people. Lower-left is better on both proxies; neither proxy measures comfort or task success.}
\Description{Four controller points: stepped B0 has the highest jerk and lowest error; interpolated B2 cuts jerk with slightly more error; jerk-limited B3 (25\,m/s$^3$ limit) cuts jerk further with more error; entropy-gated A2 is worse than B2 on both axes.}
\label{fig:frontier}
\end{figure}

\textbf{R2: lower jerk is not free (RQ2).} Figure~\ref{fig:frontier} shows the geometry on human input at 200\,ms mean delay. Stepped B0 averages 615\,m/s$^3$ at 9.91\,cm tracking error. Interpolation (B2) moves to 106\,m/s$^3$ at 11.87\,cm. Jerk-limited B3 reaches 23.5\,m/s$^3$, but at 14.79\,cm. Entropy-gated A2 (150\,m/s$^3$, 16.72\,cm) is \textbf{worse than B2 on both axes} for this input. The benchmark thus identifies a useful intervention, a costly one, and a plausible heuristic that fails.

\begin{figure}[t]
\centering\includegraphics[width=\columnwidth]{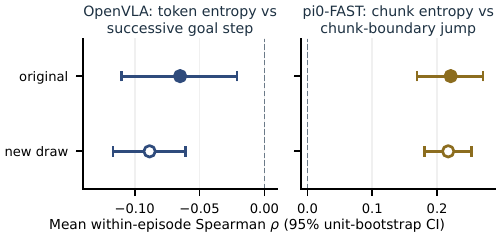}
\caption{Entropy vs command discontinuity, original (filled) and new-draw (hollow) episodes kept separate; 95\% unit-bootstrap intervals. The two panels use different action representations and discontinuity events, so their signs are not a model comparison.}
\Description{OpenVLA associations are negative in both original and new episodes; pi0-FAST associations are positive in both. No interval crosses zero.}
\label{fig:uncertainty}
\end{figure}

\begin{figure*}[t]
\centering\includegraphics[width=\textwidth]{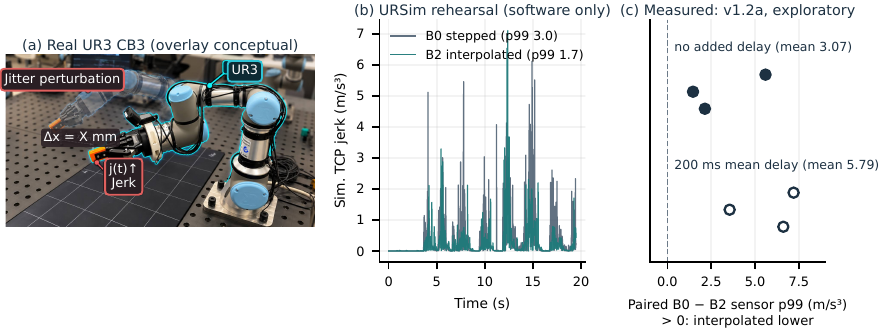}
\caption{Physical reality check. (a) The real UR3 CB3 setup; the translucent offset pose and its annotation are a \emph{conceptual illustration}. (b) Simulator rehearsal of the replay pipeline (URSim, the Universal Robots offline simulator; software path only, not hardware). (c) Measured, exploratory v1.2a session: each marker is one paired direct-mode repeat's B0$-$B2 difference in p99 tool-accelerometer-derived jerk. All six are positive; three repeats per delay are descriptive, not a significance test.}
\Description{Three panels: a UR3 photo with an illustrative translucent offset pose; a simulated jerk trace in which interpolated control has lower peaks than stepped control; six positive paired differences for no added delay and 200 ms mean delay.}
\label{fig:physical}
\end{figure*}

\textbf{R3: uncertainty is representation-dependent (RQ3; frozen, replicated).} In OpenVLA, mean translational-token entropy was weakly \emph{negatively} associated with the successive goal step (mean within-episode Spearman $\rho=-.065$ [$-.110$, $-.021$]; new 200: $-.089$ [$-.117$, $-.061$]). In pi0-FAST, chunk-token entropy was \emph{positively} associated with the chunk-boundary jump ($\rho=.221$ [$.169$, $.271$]; new 199: $.217$ [$.181$, $.253$]) (Fig.~\ref{fig:uncertainty}). \textbf{Both directions replicated.} Because the events and representations differ, the defensible conclusion is not that one model's uncertainty is better. Instead, \textbf{whether uncertainty tracks downstream discontinuity depends on the pipeline}, so uncertainty-based control needs pipeline-specific validation. Controller outcomes follow the same split. On pi0-FAST, A2 lowered p99 jerk against B2 by 208\,m/s$^3$ (new draw: 199) for +0.18\,cm (+0.16) tracking. It also beat the best attainable constant gain on feasible subsets ($-12.4$\,m/s$^3$ on 66/80; $-5.6$ on 162/199); feasibility was incomplete, and that audit was post hoc for the original set. On OpenVLA, A2 did not help: +2.3 [$-0.7$, 5.2]; new draw +3.4 [1.9, 4.8]. Table~\ref{tab:replication} summarises what replicated: all pre-registered pi0-FAST contrasts (28/28) and 22/28 for OpenVLA, whose six misses were mostly cells near zero.

\begin{table*}[t]
\caption{What replicated on new, disjoint episode draws (planned v1.4 test unless noted). Intervals are 95\% unit bootstraps.}
\label{tab:replication}\centering\footnotesize\setlength{\tabcolsep}{6pt}
\begin{tabular}{lllc}
\toprule
Finding & Original & New draw & Held?\\\midrule
OpenVLA entropy vs goal step & $-.065$ [$-.110$, $-.021$] & $-.089$ [$-.117$, $-.061$] & yes\\
pi0-FAST entropy vs chunk jump & $.221$ [$.169$, $.271$] & $.217$ [$.181$, $.253$] & yes\\
B2$-$B0 jerk, 3 VLAs & $-202$ / $-371$ / $-374$ & $-193$ / $-343$ / $-363$ & yes\\
pi0-FAST controller contrasts & 28 planned cells & 28 / 28 & yes\\
OpenVLA controller contrasts & 28 planned cells & 22 / 28 & partly\\
pi0-FAST A2 vs best constant gain$^\dagger$ & $-12.4$ (66/80) & $-5.6$ (162/199) & direction\\\bottomrule
\end{tabular}
\\[2pt]\raggedright\footnotesize $^\dagger$Feasible subsets only; post hoc for the original set. pi0.5 was added in v1.5 (post-outcome).
\end{table*}

\textbf{R4: a bounded physical check (RQ4).} The frozen v1.2 protocol replayed an author teleoperation trace in 30 small-amplitude runs on a UR3 CB3 at 125\,Hz. Position-derived p99 filtered jerk was 77.7--81.9\,m/s$^3$ across conditions, and the planned contrast was \emph{not} resolved at that logging resolution. That null result exposed a measurement limit, so an explicitly exploratory v1.2a session repeated the same motion plan while logging the robot's tool-flange accelerometer (Fig.~\ref{fig:physical}). In direct mode, \textbf{B0 exceeded B2 in all three paired repeats at each delay}, by 3.07\,m/s$^3$ on average without added delay and 5.79 with 200\,ms mean delay. B2 reductions were 3.0--14.0\%, and plant-mode differences had mixed signs. The sensor read p99 3.79 when stationary and 43.05--51.52 during motion; its gain, bandwidth and the source of the moving signal were not established. The same session's position metric also showed positive direct contrasts, so the accelerometer is not the sole source of the direction. It does not validate the point-mass plant as a UR3 model, and it is not a calibrated tool-centre-point jerk or safety result.

\section{Code structure, usage and maintenance}
The Python code has four parts: \texttt{core/} (command timing and FIFO delivery, the impedance plant, the paired stepped, interpolated and entropy-gated conditions and their metrics; NumPy and SciPy), \texttt{benchmark/} (frozen protocols, experiment drivers, paired analysis, and the v1.5 add-ons: Ruckig jerk-limited controllers, LDLJ/SPARC and trace delays), \texttt{jittervla/} (the command-line tool) and \texttt{tests/} (34 unit tests); \texttt{scripts/} holds the UR3 session and analysis tools. \texttt{python -m jittervla replay -{}-csv my.csv} replays one's own time-stamped targets and \texttt{stream -{}-npz} a stored VLA stream, each printing every condition under the same arrivals. The code will be released under Apache-2.0 after the review period, in a public repository with versioned releases and an archived snapshot linked from the project page. Issues and pull requests are reviewed by the authors, and any change to controller semantics or results is released as a new protocol version rather than silently replacing a frozen result.

\section{Availability and ethics}
The code release will contain the code (Apache-2.0), tests, a CPU demo, frozen protocols, the transformed VLA streams (CC~BY~4.0, BridgeData/DROID attribution), the 20 hand windows and the latency traces; the reported tables regenerate on a CPU without model weights or a robot. Hand trajectories are a secondary analysis of the public, de-identified 3D-ARM-Gaze dataset~\cite{lento2024armgaze} (Apache-2.0; consent and ethics approval are reported by its authors), redistributed with attribution and a statement of changes. The UR3 teleoperation episodes were recorded by an author and are released at \url{https://huggingface.co/datasets/armteam/JitterVLA-teleop-episodes}; no human-subject study was performed.

\section{Limitations and conclusion}
\begin{itemize}
\item \textbf{Model.} The main results use one generic point-mass plant with open-loop VLA predictions; closed-loop behaviour, task success and robot-specific dynamics are outside scope.
\item \textbf{Comparability.} The three pipelines differ jointly in model, dataset, action representation, entropy definition and discontinuity event, so they do not form a controlled model comparison. Service times and traces come from one GPU and one lab's network paths.
\item \textbf{Hardware.} The UR3 study is small (30 + 30 runs, one session each), the accelerometer is uncalibrated, and the v1.2a follow-up was designed after the v1.2 null result.
\item \textbf{HRI interpretation.} All metrics are motion-quality proxies. Comfort, preference, trust and perceived safety were not measured; JitterVLA identifies which conditions warrant human-centred evaluation.
\end{itemize}
\textbf{Conclusion.} Across 69,988 paired replays from synthetic reaches, 20 people and three VLA pipelines, interpolation cut modeled p99 jerk for every source (by 83\% on human motion with 200\,ms mean delay) at a 0.3--2.4\,cm tracking cost, and the effect replicated on 598 new episodes. Entropy gating helped on pi0-FAST, whose uncertainty tracked chunk jumps, but not on OpenVLA. On a real UR3, interpolation lowered accelerometer-derived jerk by 3.0--14.0\% in all six exploratory pairs. The released code and data let HRI researchers screen controller choices before human-centred studies.

\FloatBarrier
\bibliographystyle{ACM-Reference-Format}
\bibliography{refs_hri27}
\end{document}